\pdfoutput=1

\documentclass[11pt]{article}

\usepackage[final]{acl}

\usepackage{booktabs}
\usepackage{multirow}
\usepackage{hyperref}
\usepackage{multicol}
\usepackage{arydshln}
\usepackage{graphicx}
\usepackage{tabularx}
\usepackage{lscape}
\usepackage{times}
\usepackage{latexsym}
 \usepackage{longtable}
\usepackage[T1]{fontenc}

\usepackage[utf8]{inputenc}

\usepackage{microtype}

\usepackage{inconsolata}

\title{TrustAI at SemEval-2024 Task 8: A Comprehensive Analysis of Multi-domain Machine Generated Text Detection Techniques}

\author{Ashok Urlana \enspace \enspace Aditya Saibewar \enspace \enspace Bala Mallikarjunarao Garlapati \\  \enspace \enspace \textbf{Charaka Vinayak Kumar} \enspace \enspace  \textbf{Ajeet Kumar Singh} \enspace \enspace \textbf{Srinivasa Rao Chalamala}\\
TCS Research, Hyderabad, India \enspace \enspace \enspace \enspace \enspace \enspace \\
{\tt ashok.urlana@tcs.com}, {\tt aditya.saibewar@tcs.com}, {\tt balamallikarjuna.g@tcs.com}\\ {\tt charaka.v@tcs.com}, {\tt ajeetk.singh1@tcs.com}, {\tt chalamala.srao@tcs.com}
}

\begin{document}
\maketitle
\begin{abstract}
The Large Language Models (LLMs) exhibit remarkable ability to generate fluent content across a wide spectrum of user queries. However, this capability has raised concerns regarding misinformation and personal information leakage. In this paper, we present our methods for the SemEval2024 Task8, aiming to detect machine-generated text across various domains in both mono-lingual and multi-lingual contexts. Our study comprehensively analyzes various methods to detect machine-generated text, including statistical, neural, and pre-trained model approaches. We also detail our experimental setup and perform a in-depth error analysis to evaluate the effectiveness of these methods. Our methods obtain an accuracy of 86.9\% on the test set of subtask-A mono and 83.7\% for subtask-B. Furthermore, we also highlight the challenges and essential factors for consideration in future studies.
\end{abstract}
\section{Introduction}
Recent advancements in Large Language Models (LLMs) have facilitated a wide range of applications, notably in content generation \cite{Chung_2023}. While LLMs offer creative and informative content generation capabilities, concerns such as misinformation, fake news, personal information leakage, legal and ethical issues have emerged \cite{chen2023combating,li2023dark,kim2023propile}. Consequently, detecting machine-generated text has become a crucial task to address these aforementioned challenges.

The identification of machine-generated text is still an open challenge because of its overlapping similarities with human-written text.  The current text generation models produce text that is strikingly similar to human language in terms of grammaticality, coherency, fluency, and utilization of real-world knowledge \cite{radford2019language, zellers2019defending, brown2020language}. However, variations in sentence length, the presence of noisy data, and the generation of incomplete sentences are common indicators of machine-generated text.

\begin{figure}
    \centering
    \includegraphics[width=3in]{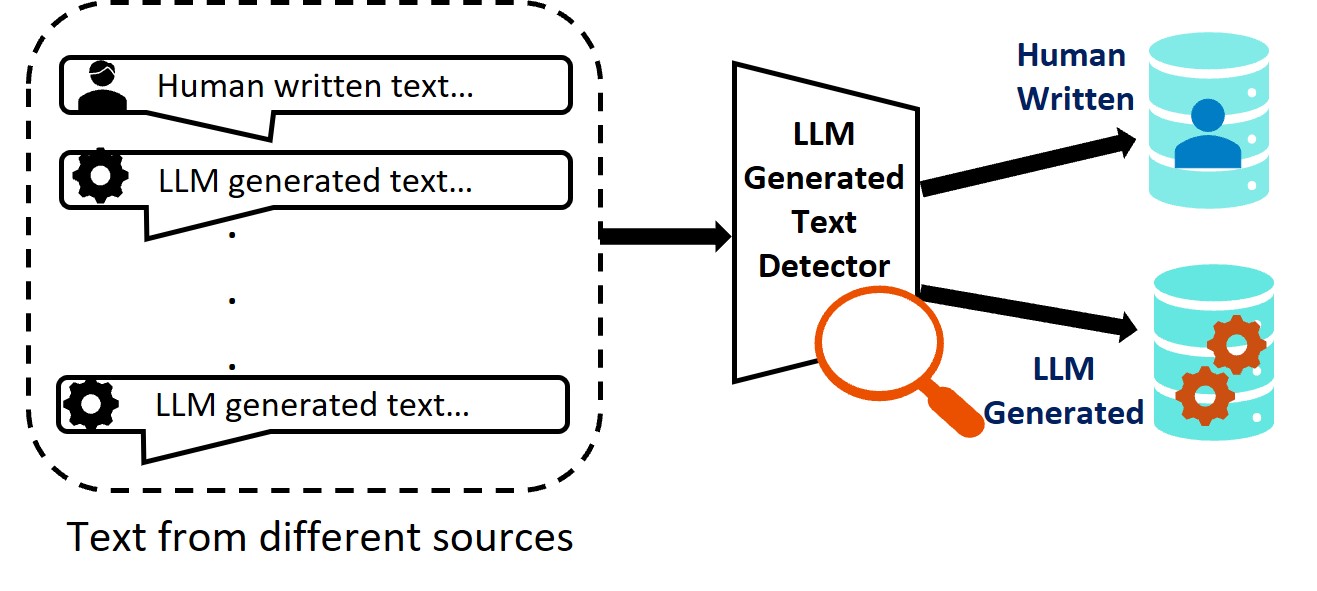}
    \caption{Block diagram for machine-generated text detection.}
    \label{fig:first_page_fig}
\end{figure}
\subsection{Essence of LLM generated text detection}
LLMs' open-ended text generation techniques have sparked various concerns across domains \cite{paper_057}. It has been demonstrated that LLMs have the potential to generate misinformation and fake news \cite{chen2023combating}, which can be catastrophic in healthcare \cite{10.1145/3544548.3581318}, public safety, education, and finance. Moreover, LLMs can generate text without source attribution, raising the risk of plagiarism \cite{quidwai-etal-2023-beyond}, and can include legal and ethical concerns too \cite{li2023dark}. 

Furthermore, when LLMs are used in enterprise applications there can be concerns of intellectual property rights infringement \cite{zhao2024silent} such as generated content might contain trademarks or branding elements \cite{ren2024copyright}. Lastly, LLMs can aggravate security concerns by generating phishing emails \cite{bethany2024large}, fake reviews \cite{10.1007/978-3-030-44041-1_114}, hallucinations \cite{huang2023survey}, biased content \cite{fang2023bias, dai2024llms}, and personal information leakage \cite{kim2023propile}.

\subsection{Tasks}
\label{sec:tasks}
The main objective of the competition is to differentiate text based on the source of its generation method (see Figure~\ref{fig:first_page_fig}), with specific importance given to machine-generated text and human-written texts \cite{semeval2024task8}. The competition consists of three tasks Subtask A, Subtask B, and Subtask C. Our study focuses on Subtasks A and B.\\
\textbf{Subtask A. Binary Human-Written vs. Machine-Generated Text Classification:} This task aims to distinguish between human-written or machine-generated text. This task acts as a binary classification. Subtask A is again subdivided into the following two categories. \textit{Mono-lingual:} The text is in the English language. \textit{Multi-lingual:} The text is in English, Chinese, Russian, Urdu, Indonesian, Arabic, and Bulgarian languages.\\
\textbf{Subtask B. Multi-Way Machine-Generated Text Classification:}
This task aims to classify the given text into six distinct classes, which are `human', `chatGPT', `cohere', `davinci', `bloomz', `dolly' with each class representing the source of its generation. This task acts as a multi-class classification.

The key contributions of this work include, 1) We present a comprehensive analysis of various machine-generated text detection techniques for multi-domain mono and multi-lingual data, 2) We provide a detailed experimental setup for statistical, neural, and pre-trained models along with corresponding error analysis, 3) We emphasize the discussions and future perspectives derived from the findings of the study.

\begin{table*}[htp]
\centering\small
\setlength{\tabcolsep}{0.6ex}
\begin{tabular}{rrcrrcrrcr}
\toprule
 & \multicolumn{3}{c}{\textbf{Subtask - A (Mono-lingual)}} & \multicolumn{3}{c}{\textbf{Subtask - A (Multi-lingual)}} & \multicolumn{3}{c}{\textbf{Subtask - B}} \\ 
\multicolumn{1}{c}{} & \multicolumn{1}{c}{Train} & \multicolumn{1}{c}{Development} & \multicolumn{1}{c}{Test} & \multicolumn{1}{c}{Train} & \multicolumn{1}{c}{Development} & \multicolumn{1}{c}{Test} & \multicolumn{1}{c}{Train} & \multicolumn{1}{c}{Development} & \multicolumn{1}{c}{Test}  \\
 \cmidrule(lr){1-1}\cmidrule(lr){2-4}\cmidrule(lr){5-7}\cmidrule(lr){8-10}
 \multicolumn{1}{l}{\# Samples} & \textbf{119757} & \phantom{0}\textbf{5000} & \textbf{34272} & \textbf{172417} &             \phantom{0}\textbf{4000} & \textbf{42378} & \textbf{71027} & \phantom{0}\textbf{3000} & \textbf{18000}\\
\multicolumn{1}{l}{\# Avg sentences} & 23 & \phantom{0}\phantom{0}\phantom{0}17 & 18 & 19 & \phantom{0}\phantom{0}\phantom{0}10& 17 & 18 & \phantom{0}\phantom{0}\phantom{0}12 & 18\\ 
\multicolumn{1}{l}{\# Minimum sentences} & 1 & \phantom{0}\phantom{0}\phantom{0}\phantom{0}1 & \phantom{0}\phantom{0}1 & 0 & \phantom{0}\phantom{0}\phantom{0}\phantom{0}1 & 1 & 1 & \phantom{0}\phantom{0}\phantom{0}\phantom{0}1 & 1\\
\multicolumn{1}{l}{\# Maximum sentences} & 1583 & \phantom{0}\phantom{0}699 & 882 & 1583 & \phantom{0}\phantom{0}\phantom{0}59 & 882 & 699 & \phantom{0}\phantom{0}477 & 882\\
\multicolumn{1}{l}{\# Median sentences} & 14 & \phantom{0}\phantom{0}\phantom{0}\phantom{0}9 & 18 & 12 & \phantom{0}\phantom{0}\phantom{0}10 & 17 & 12 & \phantom{0}\phantom{0}\phantom{0}10 & 17\\
\multicolumn{1}{l}{\# Avg words} & 530 & \phantom{0}\phantom{0}394 & 437 & 445 & \phantom{0}\phantom{0}222 & 396 & 398 & \phantom{0}\phantom{0}267 & 414\\
\multicolumn{1}{l}{\# Minimum words} & 2 & \phantom{0}\phantom{0}\phantom{0}\phantom{0}7 & 12 & 0 & \phantom{0}\phantom{0}\phantom{0}41 & 12 & 6 & \phantom{0}\phantom{0}\phantom{0}\phantom{0}7 & 12\\
\multicolumn{1}{l}{\# Maximum words} & 38070 & 19115 & 2946 & 38070 & \phantom{0}2081 & 6308 & 19115 & \phantom{0}1484 & 2946\\
\multicolumn{1}{l}{\# Median words} & 319 & \phantom{0}\phantom{0}213 & 424 & 296 & \phantom{0}\phantom{0}218 & 379 & 290 & \phantom{0}\phantom{0}217 & 413\\
 \bottomrule
\end{tabular}%
\caption{SemEval 2024 Task 8 data statistics.}
\label{tab:data_statisics}
\end{table*}

\section{Related Work}
\label{sec:related_work}

Recent works on LLM-generated\footnote{We interchangeably use the terms `LLM-generated' or `machine-generated'} text detection has shown promising results. Statistical methods are used to detect the LLM-generated text by utilizing the entropy \cite{shen2023textdefense}, and N-gram frequency \cite{9412351}. Some other studies uses the fact that language models assign high probability for the repeated sentences which is often AI model generated and ranks the AI model generated sentence  \citet{krishna-etal-2022-rankgen}. In a study, OpenAI has trained a classifier to detect LLM-generated text using the RoBERTa-based model \cite{solaiman2019release}. 

Some of the widely-used methods adopted the GPT detectors such as OpenAI detection classifier\footnote{\href{https://platform.openai.com/ai-text-classifier}{https://platform.openai.com/ai-text-classifier}}, GPTZero\footnote{\href{https://gptzero.me/}{https://gptzero.me/}}, and ZeroGPT\footnote{\href{https://www.zerogpt.com/}{https://www.zerogpt.com/}}. Another variant is DetectGPT \cite{10.5555/3618408.3619446}, which works on the assumption of LLM-generated text lies in the negative curvature region of the log-likelihood. Using this approach, DetectGPT perturbs the input text using masked language models, such as BERT \cite{DBLP:journals/corr/abs-1810-04805}, BART \cite{DBLP:journals/corr/abs-1910-13461}, T5 \cite{DBLP:journals/corr/abs-1910-10683} and compare the log probability of the text and masked filled variants. Similarly, few works utilized the different decoding strategies including top-k, nucleus, and temperature sampling to generate the text from GPT2 and BERT based models employed to perform binary classification to label text as human-written or machine generated \cite{ippolito-etal-2020-automatic}.

Recently, watermarking methods have been used in enterprises to protect the intellectual properties and fair use of the generation models. However these techniques simplify the detection of the LLM-generated output text by synonym replacement over generated outputs and text level posthoc lexical substitutions \cite{10.1145/3576915.3623120,sadasivan2023aigenerated}, and soft watermarking was introduced in \cite{pmlr-v202-kirchenbauer23a} using green and red token lists. Hidden space operations were also introduced by injecting secret signals into the probability vector of each target token \cite{10.5555/3618408.3620182}.\\
\citet{bhattacharjee2023fighting} proposed a method which triggers when the text has common words randomly assembled as it is easier to find than identifying unique and rare tokens. \citet{sadasivan2023aigenerated} focused on zero-shot AI text detection by using two clusters depending on watermarked or not. Another study \cite{wang2024mg-bench}, proposed a benchmark framework consists of an input module, a detection module and an evaluation module for machine generated text detection against human-written text. In contrast to existing works, this study presents the multi-domain multi-lingual machine generated text detection techniques.
\section{Datasets}
\label{sec:datasets}
This section given an overview of the dataset utilized and the corresponding analysis.
\subsection{Source and acquisition}
The task organizers provided the dataset\footnote{\href{https://github.com/mbzuai-nlp/SemEval2024-task8}{https://github.com/mbzuai-nlp/SemEval2024-task8}} for all the tasks (\S \ref{sec:tasks}). The dataset is an extension of the M4 dataset \cite{wang2024m4}. The dataset provided for this task consists of machine-generated text and human-written text. The human-written text is gathered from various sources such as Wikipedia, WikiHow \cite{DBLP:journals/corr/abs-1810-09305}, arXiv, and PeerRead \cite{kang-etal-2018-dataset}, Reddit \cite{fan-etal-2019-eli5} for English, Baike and Web question answering (QA) for Chinese, news for Urdu, news for Indonesian and RuATD \cite{Shamardina_2022} for Russian. On the other hand, the machine-generated text is gathered by prompting different multi-lingual LLMs: ChatGPT \cite{openai2023gpt4}, BLOOMz \cite{muennighoff-etal-2023-crosslingual}, textdavinci-003, FlanT5 \cite{chung2022scaling}, Cohere, Dolly-v2, and LLaMa \cite{touvron2023llama}. 

\subsection{Exploratory data analysis}
\label{sec:eda_data}
Preliminary analysis of data is a crucial step that is required to understand the dataset characteristics. We have observed that the number of sentences in each task data varies from 1 to a few hundred. Particularly, a few samples in the multi-lingual training data consist of empty samples as well. Another point to note is, that the number of sentences in the multi-lingual train and development varies a lot, which indicates the dataset obtained from different sources. There are a few cases, where some of the samples consist of more than 38k tokens in a single sample. With these observations, to experiment on cleaned data, we employ two types of pre-processing settings. The former (Version-1) applies heuristic-based pre-processing and sub-word removal, whereas the latter (Version-2) applies only heuristic-based pre-processing. We reported the detailed analysis of the dataset statistics in Table~\ref{tab:data_statisics}.

\section{System Overview}
\label{sec:models_description}
This section offers various approaches employed to perform machine-generated text identification. Our approaches are categorized into 1) statistical, 2) neural, and 3) pre-trained models.
\subsection{Methodology}
\subsubsection{Statistical methods}
To understand the effectiveness of statistical models, we experimented with a wide range of statistical models and their variants including ensemble approaches. The statistical models including Logistic Regression (LR), SVM, MLP, LightGBM and some of the ensemble models detailed in Table~\ref{tab:ensemble_dev_results}.
\begin{table*}[htp]
\centering\small
\setlength{\tabcolsep}{0.7ex}
\begin{tabular}{@{}lcccccccccccc@{}}
\toprule
 & \multicolumn{4}{c}{\textbf{Subtask - A (Monolingual)}} & \multicolumn{4}{c}{\textbf{Subtask - A (Multilingual)}} & \multicolumn{4}{c}{\textbf{Subtask - B}} \\ 
\multicolumn{1}{l}{\textbf{Models}} & \multicolumn{1}{c}{Count} & \multicolumn{1}{c}{Word} & \multicolumn{1}{c}{N-gram} & \multicolumn{1}{c}{Character} & \multicolumn{1}{c}{Count} & \multicolumn{1}{c}{Word} & \multicolumn{1}{c}{N-gram} & \multicolumn{1}{c}{Character} & \multicolumn{1}{c}{Count} & \multicolumn{1}{c}{Word} & \multicolumn{1}{c}{N-gram} & \multicolumn{1}{c}{Character} \\
 \cmidrule(lr){1-1}\cmidrule(lr){2-5}\cmidrule(lr){6-9}\cmidrule(lr){10-13}
\multicolumn{1}{l}{LR} & \multicolumn{1}{c}{0.544} & \multicolumn{1}{c}{0.566} & \multicolumn{1}{c}{\textbf{0.712}} & \multicolumn{1}{c}{0.615} & \multicolumn{1}{c}{0.511} & \multicolumn{1}{c}{0.516} & \multicolumn{1}{c}{0.498} & \multicolumn{1}{c}{0.561} & \multicolumn{1}{c}{0.544} & \multicolumn{1}{c}{0.514} & \multicolumn{1}{c}{0.519} & \multicolumn{1}{c}{0.558} \\ 
\multicolumn{1}{l}{Naive Bayes} & \multicolumn{1}{c}{0.506} & \multicolumn{1}{c}{0.520} & \multicolumn{1}{c}{0.568} & \multicolumn{1}{c}{0.599} & \multicolumn{1}{c}{0.510} & \multicolumn{1}{c}{0.515} & \multicolumn{1}{c}{0.489} & \multicolumn{1}{c}{0.509} & \multicolumn{1}{c}{0.463} & \multicolumn{1}{c}{0.533} & \multicolumn{1}{c}{0.495} & \multicolumn{1}{c}{0.354} \\ 
\multicolumn{1}{l}{SVM} & \multicolumn{1}{c}{0.534} & \multicolumn{1}{c}{0.573} & \multicolumn{1}{c}{0.708} & \multicolumn{1}{c}{0.634} & \multicolumn{1}{c}{0.344} & \multicolumn{1}{c}{0.494} & \multicolumn{1}{c}{0.512} & \multicolumn{1}{c}{0.571} & \multicolumn{1}{c}{0.569} & \multicolumn{1}{c}{0.550} & \multicolumn{1}{c}{0.518} & \multicolumn{1}{c}{0.573} \\ 
\multicolumn{1}{l}{Random Forest} & \multicolumn{1}{c}{0.576} & \multicolumn{1}{c}{0.614} & \multicolumn{1}{c}{0.619} & \multicolumn{1}{c}{0.682} & \multicolumn{1}{c}{0.465} & \multicolumn{1}{c}{0.517} & \multicolumn{1}{c}{0.504} & \multicolumn{1}{c}{0.559} & \multicolumn{1}{c}{0.579} & \multicolumn{1}{c}{0.462} & \multicolumn{1}{c}{0.429} & \multicolumn{1}{c}{0.408} \\ 
\multicolumn{1}{l}{XG Boost} & \multicolumn{1}{c}{0.584} & \multicolumn{1}{c}{0.623} & \multicolumn{1}{c}{0.639} & \multicolumn{1}{c}{-} & \multicolumn{1}{c}{0.499} & \multicolumn{1}{c}{0.507} & \multicolumn{1}{c}{0.558} & \multicolumn{1}{c}{-} & \multicolumn{1}{c}{0.605} & \multicolumn{1}{c}{\textbf{0.619}} & \multicolumn{1}{c}{0.591} & \multicolumn{1}{c}{-} \\ 
\multicolumn{1}{l}{MLP} & 0.594 & 0.604& 0.683 & \multicolumn{1}{c}{0.647} & 0.544 & 0.528 & 0.485 & \multicolumn{1}{c}{\textbf{0.609}} & 0.529 & 0.506 & 0.493 & 0.583 \\ \bottomrule
\end{tabular}%
\caption{Accuracy of statistical models development set; LR refers to Logistic Regression, Subtask-B deals with multi-class classification task.}
\vspace{-0.9mm}
\label{tab:statistical_dev_results}
\end{table*}
\begin{table}
\centering\footnotesize
\setlength{\tabcolsep}{0.2ex}
\begin{tabular}{lcc}
\toprule
\multicolumn{1}{c}{\textbf{Model}}                             & \textbf{ \begin{tabular}[c]{@{}l@{}} Subtask-A\\(Monolingual) \end{tabular} }  & \textbf{Subtask-B} \\ \midrule
\begin{tabular}[c]{@{}l@{}}Naive Bayes + SGDClassifier + \\LightGBM \end{tabular} & 0.714 & 0.708    \\ \bottomrule
\end{tabular}%
\caption{Ensemble model Accuracy scores on development set.}
\label{tab:ensemble_dev_results}
\end{table}

\subsubsection{Neural methods}
Neural networks have demonstrated remarkable success in various domains, from image and speech recognition to natural language processing. We experiment with Convolutional Neural Networks (CNN), Recurrent Neural Networks (RNN), Long Short-term Memory (LSTM) \cite{10.1162/neco.1997.9.8.1735} and their combinations. We utilize FastText [\cite{joulin2016fasttextzip}, \cite{bojanowski2017enriching}] embeddings to capture hierarchical patterns within the text data.\\

\subsubsection{Pre-trained models} Self-supervised pre-trained models have been effective for the classification tasks. In this study, we experiment with a wide range of pre-trained models trained on either open-source or language model-generated data. The pretrained models including BERT \cite{DBLP:journals/corr/abs-1810-04805}, RoBERTa \cite{DBLP:journals/corr/abs-1907-11692}, DistilRoBERTa base \cite{Sanh2019DistilBERTAD}, RoBERTa Base OpenAI Detector \cite{solaiman2019release}, XLM RoBERTa \cite{DBLP:journals/corr/abs-1911-02116}.  \\
\subsection{Experimental setup}
For all the experiments, we have utilized the default data splits provided by the task organizers. For all the statistical models, four types of embeddings were employed namely counter vectors, word, n-gram, character-level TF-IDF vectors and spaCy embeddings. Moreover, we used the default configurations mentioned in the scikit-learn\footnote{\url{https://scikit-learn.org/stable/supervised_learning.html}}. Whereas for pre-trained models the list of hyperparameters details are listed in Table~\ref{tab:pretrain_models_experimental_setup}. We have not performed any hyperparameter-tuning for our experiments. We conduct most of our experiments using four Nvidia GeForce RTX 2080 Ti (11GB) GPUs. To evaluate all the models, we reported the `Accuracy' scores. 

\section{Results and Analysis}
\label{sec:results_analysis}
This section provides a detailed analysis of the models utilized for subtasks A and B. Our experiments aim to showcase the effectiveness of several machine-generated text detection techniques. 

\begin{table}
\centering\small
\setlength{\tabcolsep}{0.6ex}
\begin{tabular}{lccc}
\toprule
\multicolumn{1}{l}{} & \multicolumn{2}{c}{\textbf{Subtask-A}}                               & \multirow{2}{*}{\textbf{Subtask-B}} \\ 
\textbf{Model}       & \multicolumn{1}{l}{\textbf{Mono}} & \multicolumn{1}{l}{\textbf{Multi}} &                                     \\ \midrule
CNN + FastText               & 0.711 & 0.545 & 0.652 \\
RNN + LSTM + FastText        & 0.682 & 0.615 & 0.549 \\
Bidirectional RNN + FastText & 0.689 & 0.579 & 0.582 \\ \bottomrule
\end{tabular}%
\caption{Accuracy of neural models on development set.}
\label{tab:neural_dev_scores}
\end{table}
\begin{table}
\centering\small
\setlength{\tabcolsep}{0.2ex}
\begin{tabular}{lcc}
\toprule
\textbf{Task}              & \textbf{Model}                        & \textbf{Accuracy} \\ \midrule
\multirow{5}{*}{Subtask-A (Mono)}      & BERT Base            & \textbf{0.825}    \\
                           & BERT Base\_v1        & 0.807   \\
                           & BERT Base\_v2        & 0.813   \\
                           & BERT Base\_v2        & 0.809   \\
                           & RoBERTa Base OpenAI Detector & 0.766   \\ \midrule
\multirow{3}{*}{Subtask-A (Multi)}     & BERT Multilingual Base\_v2         & 0.622   \\
                           & XLM-RoBERTa             & \textbf{0.766 }  \\
                           & BERT Multilingual Base             & 0.622   \\ \midrule
\multirow{3}{*}{Subtask B} & RoBERTa Large                & 0.751  \\
                           & RoBERTa Base OpenAI Detector & \textbf{0.753 }  \\
                           & DistilRoBERTa Base          & 0.733   \\ \bottomrule
\end{tabular}%
\caption{Pre-trained models Accuracy scores on development set; Where v1 and v2 indicates different pre-processing strategies.}
\label{tab:pretrain_dev_scores}
\end{table}
\subsection{Subtask A Mono-lingual}
We experiment with the statistical and neural models to perform subtasks A and B. All the statistical and ensemble models experimental results on development data are mentioned in Table~\ref{tab:statistical_dev_results} and Table~\ref{tab:ensemble_dev_results}. The results on test data mentioned in Table~\ref{tab:test_results}. In the case of statistical models, Logistic Regression obtains the superior performance of 71.2\% accuracy using n-gram level TF-IDF embeddings compared to other methods on the development dataset. Whereas in the case of the performance of the test set, our ensemble surpass all the remaining models. We built the ensemble model by creating a custom tokenizer by combining spaCy embedding and TF-IDF with n-gram level range of (3-5) embedding. Moreover, we trained an ensemble model with Naive Bayes, SGDClassifier\footnote{\url{https://scikit-learn.org/stable/modules/generated/sklearn.linear_model.SGDClassifier.html}}, and LightGBM models which gave 86.9\% accuracy on the test set.\\
We experiment with a few neural models with fastText embeddings and out of them CNN+fastText outperforms the other models. We have listed results in Table~\ref{tab:neural_dev_scores}. Moving ahead, we fine-tuned transformer-based pre-trained language models like RoBERTa Base OpenAI detector \cite{solaiman2019release}, which gave 76.6\% accuracy on the development set and 78.7\% accuracy on test set, BERT base model which gave 82.5\% accuracy on the development set and 71.7 \% accuracy on test set. The results are detailed in Table~\ref{tab:pretrain_dev_scores}. Furthermore, we use the pre-processing steps discussed in Section~\ref{sec:eda_data}. Fine-tuned the BERT base model with version-1's pre-processed data gave 80.7\% on the development dataset and 71.7\% on the test set. Then we fine-tuned the BERT base model with version-2 pre-processed data gave 81.3\% on the development dataset and 69.7\% on the test set. We secured 
24$^{th}$ rank out of 137 participants.\\
We observed that statistical models that performed modestly on the development set generalized effectively to the test set, whereas some pre-trained language models, despite performing well on the development set, struggled to generalize on test set. This discrepancy may stem from the differing sources of the training and development sets (`arxiv', `reddit', `wikihow', `wikipedia', `peerread') compared to the test set, potentially causing over-fitting of the pre-trained models on the training data and hindering their performance on the test set.
\begin{table}
\centering\small
\setlength{\tabcolsep}{0.5ex}
\begin{tabular}{lcccccccccccc}
\toprule
\multicolumn{1}{c}{\textbf{Model}} & \multicolumn{1}{c}{\textbf{Batch size}} & \multicolumn{1}{c}{\textbf{Epochs}} & \multicolumn{1}{c}{\textbf{Vocab size}} \\
 \cmidrule(lr){1-1}\cmidrule(lr){2-4}
 \multicolumn{1}{l}{BERT Base} & \multicolumn{1}{c}{16} & \multicolumn{1}{c}{10} & \multicolumn{1}{c}{\phantom{0}30522} \\
 \multicolumn{1}{l}{OpenAI Detector} & \multicolumn{1}{c}{16} & \multicolumn{1}{c}{10} & \multicolumn{1}{c}{\phantom{0}50265} \\
 \multicolumn{1}{l}{BERT Multilingual Base}  & \multicolumn{1}{c}{\phantom{0}8} & \multicolumn{1}{c}{\phantom{0}3} & \multicolumn{1}{c}{\phantom{0}30522} \\
 \multicolumn{1}{l}{XLM-RoBERTa}  & \multicolumn{1}{c}{\phantom{0}8} & \multicolumn{1}{c}{\phantom{0}5} & \multicolumn{1}{c}{250002} \\
 \multicolumn{1}{l}{RoBERTa Large}  & \multicolumn{1}{c}{\phantom{0}4} & \multicolumn{1}{c}{\phantom{0}2} & \multicolumn{1}{c}{\phantom{0}50265} \\
 \multicolumn{1}{l}{DistilRoBERTa Base} & \multicolumn{1}{c}{16} & \multicolumn{1}{c}{10} & \multicolumn{1}{c}{\phantom{0}29409} \\
 \bottomrule
\end{tabular}%
\caption{Experimental setup for pre-trained models. For all the models max source length set to 512 and learning rate $5\mathrm{e}^{-5}$.}
\label{tab:pretrain_models_experimental_setup}
\vspace{-0.7cm}
\end{table}
\subsection{Subtask A Multi-lingual}
For subtask A multi-lingual, we fine-tuned BERT Multilingual Base and XLM RoBERTa base models. BERT Multilingual Base along with version-2 pre-processed data resulted in 62.2\% accuracy on the development set and 73.8\% accuracy on the test set. Moreover, despite the decent performance of XLM-RoBERTa on the development set with 76.6\% accuracy, the performance of on test set is sub-par. Furthermore, the BERT Multilingual base gave 62.2\% accuracy on the development set and 73.1\% accuracy on the test set. As mentioned in Section~\ref{sec:eda_data}, we observed that, the multi-lingual data consists of empty samples. Hence, we fine-tuned the BERT Multilingual Base model on the version-2 of the pre-processed data, which helped in improving the accuracy of the test set even if we had the same accuracy on development set.
\subsection{Subtask B}
Subtask B deals with multi-class classification task. For this task, we have conducted experiments using the statistical models as well as the pre-trained language models. MLP model gave the best accuracy on the development set with 60.9\% accuracy. Our ensemble approach obtains 70.8\% accuracy on the development set and 65\% accuracy on the test set.
Moreover, we experimented with RoBERTa Base OpenAI Detector gave 75.3\% on the development set and 83.7\% accuracy on the test set. Whereas, the DistilRoBERTa base obtains 73.3\% accuracy on the development set and 79.1\% accuracy on the test set and secured 17$^{th}$ rank out of 86 participants.
\begin{table}
\centering\small
\setlength{\tabcolsep}{0.2ex}
\begin{tabular}{ccc}
\toprule
\textbf{Task}              & \textbf{Model}                                        & \textbf{Accuracy} \\ \midrule
\multirow{6}{*}{\begin{tabular}[c]{@{}l@{}} Subtask - A \\ (Mono) \end{tabular}}      
                        & Baseline &   \phantom{0}0.74   \\
                        & \begin{tabular}[c]{@{}l@{}} Naive bayes + SGDClassifier \\ + LightGBM* \end{tabular} & \textbf{0.869}             \\
                           & \begin{tabular}[c]{@{}l@{}} RoBERTa Base \\ OpenAI Detector \end{tabular}                          & 0.787             \\
                           & BERT Base\_v1                                 & 0.717             \\
                           & BERT Base                                     & 0.715             \\
                           & BERT Base\_v2                                 & 0.697             \\  \midrule
\multirow{2}{*}{\begin{tabular}[c]{@{}l@{}} Subtask - A \\ (Multi) \end{tabular}}     
                            & Baseline     & \phantom{0}0.72 \\
                           & BERT Multilingual Base\_v2                                  & \textbf{0.738}             \\
                           & BERT Multilingual Base                                      & 0.731             \\
                           & XLM-RoBERTa *                                     & \phantom{0}0.50             \\ \midrule
\multirow{3}{*}{Subtask - B} 
                           & Baseline & \phantom{0}0.75 \\
                           & \begin{tabular}[c]{@{}l@{}} RoBERTa Base \\ OpenAI Detector \end{tabular}                          & \textbf{0.837}             \\
                           & DistilRoBERTa Base*                                   & 0.791             \\
                           & \begin{tabular}[c]{@{}l@{}} Naive bayes + SGDClassifier+ \\ LightGBM \end{tabular}& 0.650            \\ \bottomrule
\end{tabular}%
\caption{Test set accuracy results; *entries are the official submission models of the competition.}
\label{tab:test_results}
\end{table}


\section{Conclusions}
The study explores different methodologies for detecting machine-generation text, leveraging statistical, neural, and pre-trained models. We observe that the ensemble models are more effective in classifying the mono-lingual data (Subtask-A mono), while models trained on GPT2-text surpass other models in multi-class classification. 
\section{Limitations}
 In our study, due to computational constraints, we have not performed experiments with any large language models. Current evaluation has been limited to conventional ML and pre-tained language models. Some of our experimental methods perform better on development data, where as there is a significant drop on test data, this may result in lack of generalization.

\bibliography{custom}




\end{document}